\documentclass[10pt,twocolumn,letterpaper]{article}

\usepackage{iccv}
\usepackage{times}
\usepackage{epsfig}
\usepackage{graphicx}
\usepackage{amsmath}
\usepackage{amssymb}
\usepackage{multirow}

\usepackage[accsupp]{axessibility}
\usepackage{microtype}

\usepackage[pagebackref=true,breaklinks=true,colorlinks,bookmarks=false]{hyperref}

\newcommand{\myhash}{\raisebox{0.5\depth}{\#}}

\iccvfinalcopy 

\def\iccvPaperID{3129} 

\ificcvfinal\fi

\begin{document}

\title{A skeletonization algorithm for gradient-based optimization}

\author{Martin J. Menten$^{1,2}$ \quad Johannes C. Paetzold$^{1,2}$ \quad Veronika A. Zimmer$^{1}$ \quad Suprosanna Shit$^{1}$\\
Ivan Ezhov$^{1}$ \quad Robbie Holland$^{2}$ \quad Monika Probst$^{1}$ \quad Julia A. Schnabel$^{1}$ \quad Daniel Rueckert$^{1,2}$\\
$^{1}$Technical University of Munich \quad $^{2}$Imperial College London\\
}

\maketitle
\ificcvfinal\thispagestyle{empty}\fi

\begin{abstract}

The skeleton of a digital image is a compact representation of its topology, geometry, and scale. It has utility in many computer vision applications, such as image description, segmentation, and registration. However, skeletonization has only seen limited use in contemporary deep learning solutions. Most existing skeletonization algorithms are not differentiable, making it impossible to integrate them with gradient-based optimization. Compatible algorithms based on morphological operations and neural networks have been proposed, but their results often deviate from the geometry and topology of the true medial axis. This work introduces the first three-dimensional skeletonization algorithm that is both compatible with gradient-based optimization and preserves an object's topology. Our method is exclusively based on matrix additions and multiplications, convolutional operations, basic non-linear functions, and sampling from a uniform probability distribution, allowing it to be easily implemented in any major deep learning library. In benchmarking experiments, we prove the advantages of our skeletonization algorithm compared to non-differentiable, morphological, and neural-network-based baselines. Finally, we demonstrate the utility of our algorithm by integrating it with two medical image processing applications that use gradient-based optimization: deep-learning-based blood vessel segmentation, and multimodal registration of the mandible in computed tomography and magnetic resonance images.
\end{abstract}

\section{Introduction}

\begin{figure*}[t!]
\centering
\includegraphics[width=0.875\linewidth]{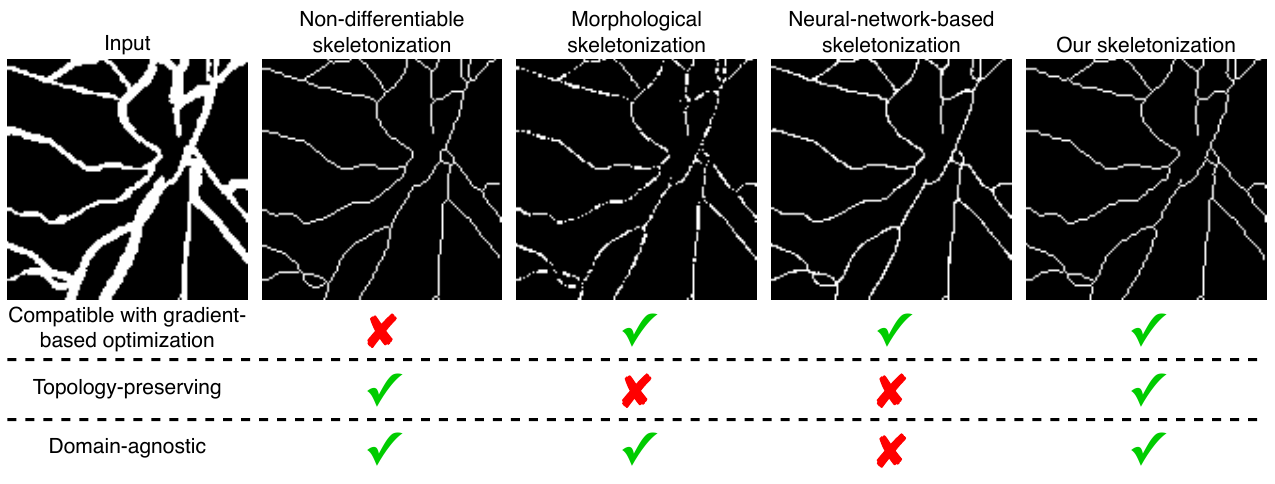}
\caption{Most existing skeletonization algorithms are not differentiable, making it impossible to integrate them with gradient-based optimization. Morphological and neural-network-based solutions can be used with backpropagation, but alter the topology of the object by introducing breaks along the skeleton. Our proposed skeletonization algorithm preserves the topology while simultaneously being compatible with gradient-based optimization.}
\label{fig:graphical_abstract}
\end{figure*}

Skeletonization algorithms aim at extracting the medial axis of an object, which is defined as the set of points that have more than one closest point on the object's boundary \cite{blum1967transformation}. This lower-dimensional representation compactly encodes various geometric, topological, and scale features. As such, it is useful for many tasks in computer vision, including object description, compression, recognition, tracking, registration, and segmentation \cite{ma1996fully,morse1993multiscale,palagyi2001sequential,thibault2000terrain,zhao2007preprocessing}. While there are several efficient approaches to calculate the medial axis in continuous space, extracting the skeleton of a discrete digital image is not trivial. The search for accurate skeletonization algorithms, which can process two-dimensional and three-dimensional digital images, has spawned a plethora of research works \cite{bertrand1995three,bertrand2014powerful,borgefors1999computing,lee1994building,ma1996fully,saha1997new,tsao1981parallel,zhou1999efficient}. For a comprehensive overview and taxonomy of skeletonization algorithms and their applications, we refer to the excellent survey by Saha \textit{et al.} \cite{saha2016survey}.

Today, computer vision tasks are commonly solved using deep learning. Skeletonization may be used as building block or inductive bias in these image processing applications \cite{jerripothula2017object,liu2020dynamic,shit2021cldice}. However, most established skeletonization methods are not compatible with backpropagation and gradient-based optimization \cite{saha2016survey}. The few works that have integrated skeletonization with deep learning pipelines rely on morphological skeletonization algorithms \cite{shit2021cldice}. This class of algorithms is based on simple morphological operations, such as erosion and dilation \cite{maragos1986morphological,viti2022coronary}. However, they will often result in breaks in the skeleton, causing it to diverge from the geometry and topology of the true medial axis. Recently, learning-based methods have also been harnessed for skeletonization \cite{demir2019skelneton,jerripothula2017object,liu2020dynamic,nathan2019skeletonnet,nguyen2021u,panichev2019u,shen2016object,shen2017deepskeleton}. Most of these works train an encoder-decoder neural network on pairs of input images and ground truth skeletons, which have previously been obtained using a classical skeletonization algorithm. While learned approaches are intrinsically compatible with backpropagation, they are not guaranteed to preserve the topology of the input. Additionally, they are susceptible to domain shifts between the training and inference data \cite{ganin2016domain,vazquez2013virtual}.

\noindent\textbf{Our contribution} This work bridges the gap between traditional skeletonization principles with strict topology guarantees and their integration with gradient-based optimization. We introduce a skeletonization algorithm that is topology-preserving, domain-agnostic, and compatible with backpropagation (see Figure \ref{fig:graphical_abstract}). Our algorithm is exclusively based on matrix additions and multiplications, convolutional operations, basic non-linear functions, and sampling from a uniform probability distribution, allowing it to be easily implemented in any major deep learning library, such as PyTorch or Tensorflow \cite{tensorflow2015-whitepaper,paszke2017automatic}. In benchmarking experiments, we establish that our algorithm outperforms non-differentiable, morphological, and neural-network-based baselines. Finally, we directly integrate our skeletonization algorithm with two medical image processing applications that rely on gradient-based optimization. We show that it enhances both deep-learning-based blood vessel segmentation, and multimodal registration of the mandible in computed tomography (CT) and magnetic resonance (MR) images.

\section{Prerequisites}

\noindent\textbf{Digital images} A discrete three-dimensional image is an array of points $P = \{p(x, y, z)\}$, which are each assigned an intensity value, on a lattice defined by Cartesian coordinates $x, y, z \in \mathbb{Z}$ \cite{kong1989digital}. Owing to the grid nature of the image, we can define the 6-neighborhood $N_6(p)$, 18-neighborhood $N_{18}(p)$, and 26-neighborhood $N_{26}(p)$ of a point \cite{rosenfeld1976digital}:
\thinmuskip=1mu
\medmuskip=2mu
\thickmuskip=4.5mu
\begin{align}
\hspace{-1cm}N_6(p) &= \{p'; (|x - x'| + |y - y'| + |z - z'|) \leq 1\}\,,\notag\\
N_{26}(p) &= \{p'; \max(|x - x'|, |y - y'|, |z - z'|) \leq 1\}\,,\\
N_{18}(p) &= \{p'; (|x - x'| + |y - y'| + |z - z'|) \leq 2\} \cap N_{26}(p)\,\notag.
\end{align}
\thinmuskip=3mu 
\medmuskip=4mu 
\thickmuskip=5mu 
\noindent
Points that are inside each other's $n$-neighborhood are called $n$-adjacent. Two points $p$ and $p'$ are said to be $n$-connected, if there is a sequence of points $p = p_0, ..., p_k = p'$ so that each $p_i$ is $n$-adjacent to $p_{i - 1}$ for $1 \leq i \leq k$.

\noindent\textbf{Euler characteristic} In the special case of a binary image the value of each point is either 1 or 0. The foreground of a binary image is represented by the set $S$ of points with value 1, while the set $\overline{S}$ denotes the remaining points with value 0. Based on above's definition of connectedness, we can define an object $O$ as a set of $n$-connected points in $S$. An object in $\overline{S}$ that is completely surrounded by points in $S$ is called a cavity $C$ in $S$. Finally, a hole $H$ can be intuitively described as a tunnel through $S$. Objects, cavities and holes can be analogously defined for $\overline{S}$. By combining the number of objects, cavities, and holes the Euler characteristic, or genus, of a 6-connected set of points $G_6$ can be determined \cite{kong1989digital}:
\begin{equation}
\label{eq:g6}
G_6 = \myhash O - \myhash H + \myhash C\,.
\end{equation}
\noindent
It is also possible to span a graph between all 6-neighbors. On this graph, simplicial complexes consisting of one, two, four, or eight points are called vertex $v$, edge $e$, face $f$, or octant $oct$, respectively \cite{kong1989digital}. $G_6$ can also be calculated based on the number of these complexes via
\begin{equation}
\label{eq:g6_simplical}
G_6 = \myhash v - \myhash e + \myhash f - \myhash oct\,.
\end{equation}
\noindent
In the following, we only consider the case in which objects in $S$ are 26-connected and objects in $\overline{S}$ are 6-connected. To calculate the genus of a 26-connected $S$ $G_{26}(S)$ we can derive $G_6(\overline{S})$ using Equation \ref{eq:g6} or \ref{eq:g6_simplical} and use the following relation \cite{kong1989digital}:
\begin{equation}
\label{eq:genus_relation}
G_{26}(S) = G_6(\overline{S}) - 1
\end{equation}

\noindent\textbf{Simple points} Crucial to the skeletonization of digital images is the definition of a simple point. A point belonging to $S$ is simple if it can be deleted, that is changed from 1 to 0, without altering the image's topology. Morgenthaler \cite{morgenthaler1981three} shows that this is the case if the deletion of a point does not change the number of objects and holes in $S$ and $\overline{S}$. Using $\delta$ to denote the difference in topology between $S$ and $S \setminus \{p\}$, we can write this relation as:
\begin{equation}
\begin{split}
p &\textrm{ is simple} \iff \\
&\delta O(S) = 0\,,\delta O(\overline{S}) = 0\,,\delta H(S) = 0\,,\delta H(\overline{S}) = 0\,.
\end{split}
\end{equation}
\noindent
Lee \etal \cite{lee1994building} prove that these conditions are equivalent to calculating the change in the number of objects and Euler characteristic of $S$ in a point's local 26-neighborhood:
\begin{equation}
\label{eq:lee_criteria}
p \textrm{ is simple} \iff \delta O(S) = 0\,,\delta G_{26}(S) = 0\,.
\end{equation}

\section{Method}

\begin{figure}[h!]
\centering
\includegraphics[width=0.95\linewidth]{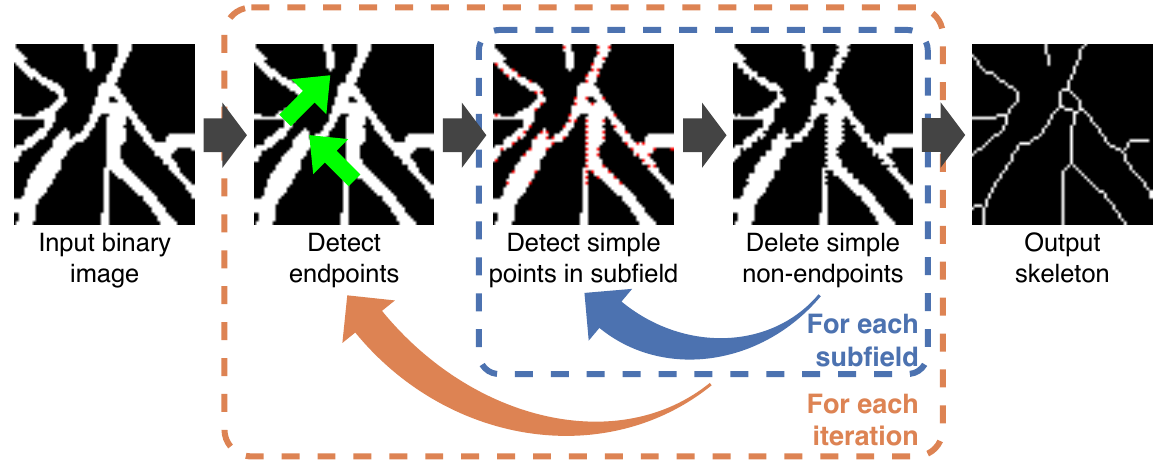}
\caption{Data flow through an iterative boundary-peeling skeletonization algorithm. Our method follows the same paradigm, while ensuring that the identification of simple points, endpoints, and the subfield-based parallelization are all compatible with gradient-based optimization.}
\label{fig:algorithm_flowchart}
\end{figure}

Arguably, the most common class of skeletonization algorithms for digital images are iterative boundary-peeling methods \cite{saha2016survey}. These algorithms are based on the repeated removal of simple points until only the skeleton remains. \cite{bertrand1996boolean,lee1994building,lobregt1980three,morgenthaler1981three}. At its core, our skeletonization algorithm follows the same paradigm (see Figure \ref{fig:algorithm_flowchart}). To ensure that our method is compatible with gradient-based optimization while remaining topology-preserving, we make the following contributions:
\begin{itemize}
\setlength\itemsep{0em}
\item We introduce two methods to differentiably identify simple points (see Section \ref{sec:simple_point_identification}). One solution relies on the calculation of the Euler characteristic, and the other one is based on a set of Boolean rules that evaluate a point's 26-neighborhood.
\item We adopt a scheme to safely delete multiple simple points at once, enabling the parallelization of our algorithm (see Section \ref{sec:parallelization}).
\item We introduce a strategy to apply our algorithm to non-binary inputs and integrate it with gradient-based optimization by employing the reparametrization trick and a straight-through estimator (see Section \ref{sec:continuous_inputs}).
\item All of above's contributions are formulated using matrix additions and multiplications, convolutional operations, basic non-linear functions, and sampling from a uniform probability distribution. We combine them into a single PyTorch module, which we make publicly available (see Section \ref{sec:implementation}).
\end{itemize}

\subsection{Identification of simple points}
\label{sec:simple_point_identification}

\subsubsection{Euler characteristic to identify simple points}

Lobregt \etal \cite{lobregt1980three} base their detection of simple points on the observation that the removal of a simple point does not alter the genus of its 26-neighborhood:
\begin{equation}
\label{eq:lobregt}
p \textrm{ is simple} \implies \delta G_{26}(S) = 0\,,
\end{equation}
which is a relaxation of Equation \ref{eq:lee_criteria}. To efficiently determine $\delta G_{26}$(S), their algorithm assesses the change of the genus in each of the 26-neighborhood's eight octants and sums their contributions. Thereby, they rely on a look-up table in which each entry corresponds to one of the $2^8$ possible configurations of an octant.

\begin{figure*}[h!]
\centering
\includegraphics[width=0.95\linewidth]{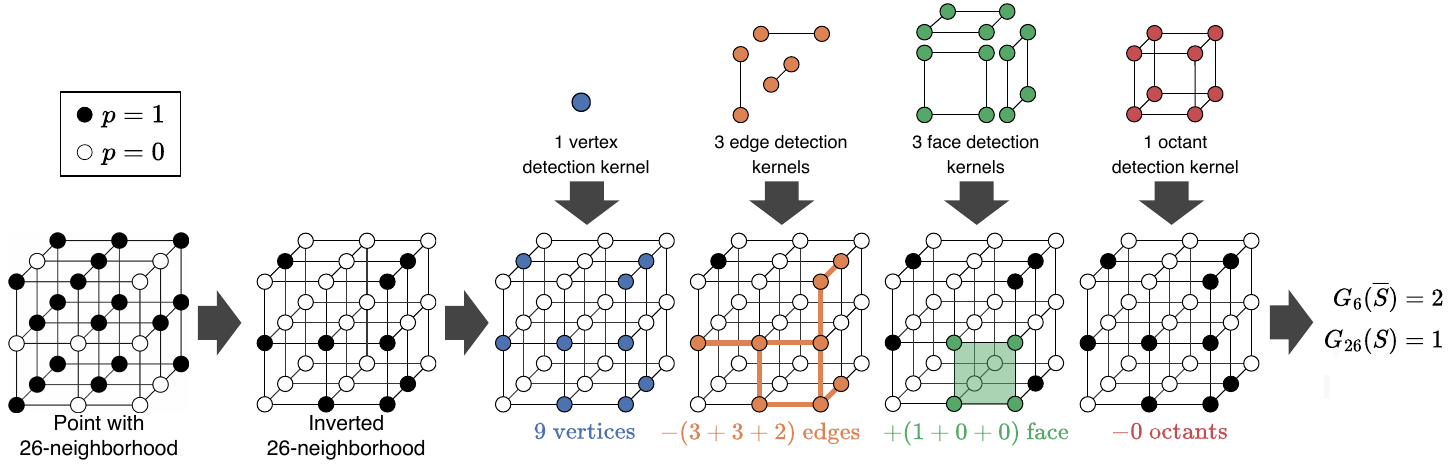}
\caption{In order to calculate the Euler characteristic of a point, we initially invert its 26-neighborhood. Next, we determine the number of vertices, edges, faces, and octants of the background via eight simple convolutions with predefined kernels. Finally, Equations \ref{eq:g6_simplical} and \ref{eq:genus_relation} are used to derive the Euler characteristic of the foreground.}
\label{fig:genus_calculation}
\end{figure*}

In order to reduce the number of comparisons and provide a smoother learning signal for backpropagation, we use eight convolutions with pre-defined kernels to determine the number of vertices, edges, faces, and octants (see Figure \ref{fig:genus_calculation}). By inserting these into Equation \ref{eq:g6_simplical}, we calculate $G_6(\overline{S})$ of each 26-neighborhood. Afterwards, we repeat this process with the central point of each neighborhood set to zero and assess whether the Euler characteristic has changed. This process is parallelized while ensuring that only one point in each 26-neighborhood is deleted at a given time. To this end, we use multiple sets of points given by
\begin{equation}
\label{eq:subfields}
\begin{split}
S_{i,j,k} \in \,&\{p'(x + i, y + j, z + k)\};\\
&x,y,z\in\{0,2,4,\ldots\},\; i,j,k\in\{0,1\}\,
\end{split}
\end{equation}
Cycling through all combinations of $i$, $j$, and $k$ yields eight subfields of points that can be processed simultaneously. The same subfields are also used during the later removal of simple points (see Section \ref{sec:parallelization}).

Lee \etal \cite{lee1994building} show that the invariance of the Euler characteristic under deletion of a point is a necessary but not sufficient condition for it being simple (cf. Equation \ref{eq:lee_criteria}). Consequently, above's strategy slightly overestimates the number of simple points \cite{lee1994building}. On the set of all possible $2^{26}$ configurations of a 26-neighborhood, above's algorithm characterizes the central point as simple in 40.07\% of cases when in fact only 38.72\% are truly simple.

\subsubsection{Boolean characterization of simple points}

\begin{figure}[h!]
\centering
\includegraphics[width=0.9\linewidth]{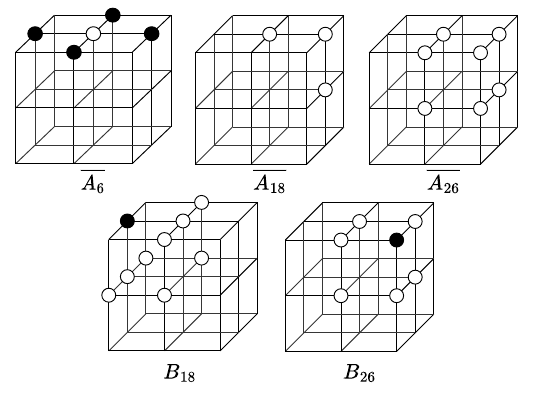}
\caption{The five cell configuration introduced by Bertrand \etal \cite{bertrand1996boolean} used for Boolean characterization of simple points.}
\label{fig:bertrand_cell_configurations}
\end{figure}

For this reason, we propose a second method that identifies the exact set of simple points. It is based on work by Bertrand \etal \cite{bertrand1996boolean} who introduce the following Boolean characterization of a simple point:
\begin{equation}
\label{eq:bertrand}
\begin{split}
p &\textrm{ is simple} \iff (\myhash\overline{X_6} = 1) {\;} or {\;} (\myhash X_{26} = 1) {\;}\\
& or {\;} (\myhash B_{26} = 0, \myhash X_{18} = 1) \\
& or {\;} (\myhash \overline{A_6} = 0, \myhash B_{26} = 0,\myhash B_{18} = 0,\\
& {\;} {\;} {\;} {\;} {\;} {\;} \myhash \overline{X_6} - \myhash \overline{A_{18}} + \myhash \overline{A_{26}} = 1) \,
\end{split}
\end{equation}
where $\myhash X_n$ and $\myhash \overline{X_n}$ are the number of $n$-neighbors of a point belonging to $S$ and $\overline{S}$, respectively. $\myhash B_{26}$, $\myhash \overline{A_6}$, $\myhash B_{18}$, $\myhash \overline{A_{18}}$, and $\myhash \overline{A_{26}}$ correspond to the number of specific cell configurations depicted in Figure \ref{fig:bertrand_cell_configurations}.

Similar as before, the presence of these five configurations and their 6-, 8-, and 12-rotational equivalents can be efficiently checked by convolving the image with pre-defined kernels. Compared to our first strategy, this algorithm trades off computational efficiency for accuracy. It requires a total of 57 convolutions with three-dimensional kernels, but is guaranteed to correctly classify all points of a binary image as either simple or not.

\subsection{Parallelization and endpoint conditions}
\label{sec:parallelization}

Sequentially checking each point using above's conditions and deleting them if they are simple already constitutes a functioning skeletonization algorithm. However, this strategy is very inefficient when applied to large three-dimensional images. Naively deleting all simple points at once is not possible as simultaneous removal of neighboring points may affect the object's topology even if both points are simple. For this reason, previous works have researched strategies to safely remove multiple simple points in parallel \cite{bertrand1995three,lee1994building,ma1996fully,morgenthaler1981three,saha1997new,tsao1981parallel}. We adopt a subiterative scheme based on the same eight subfields that are already used during the calculation of the Euler characteristic (see Equation \ref{eq:subfields}) \cite{bertrand1995three}. In conventional skeletonization algorithms, the program terminates once a full iteration does not result in any points being deleted. To keep the number of operations comparable during repeated application, we explicitly provide the number of outer-loop iterations. This simple scalar hyperparameter can be easily tuned on a few representative samples of the considered dataset.

Merely preserving non-simple points would lead to a topological skeleton. For example, a solid object without any holes or cavities would be reduced to a single point. For many applications in image processing, it is desirable to also preserve some information about the image's geometry, such as the existence and position of extremities. This can be achieved by also preserving so-called endpoints. Our algorithm uses the following definition of an endpoint:
\begin{equation}
p\textrm{ is endpoint } \iff \myhash X_{26} \leq 1
\end{equation}
Other definitions for endpoints could potentially be integrated with our algorithm and would result in different properties of the obtained skeleton. For example, endpoint conditions could be chosen to extract a medial surface instead of a medial axis, or the number of short extremities, sometimes called spurs, may be reduced \cite{shaked1998pruning,zhou1999efficient}.

\begin{figure*}[ht!]
\centering
\includegraphics[width=1.0\linewidth]{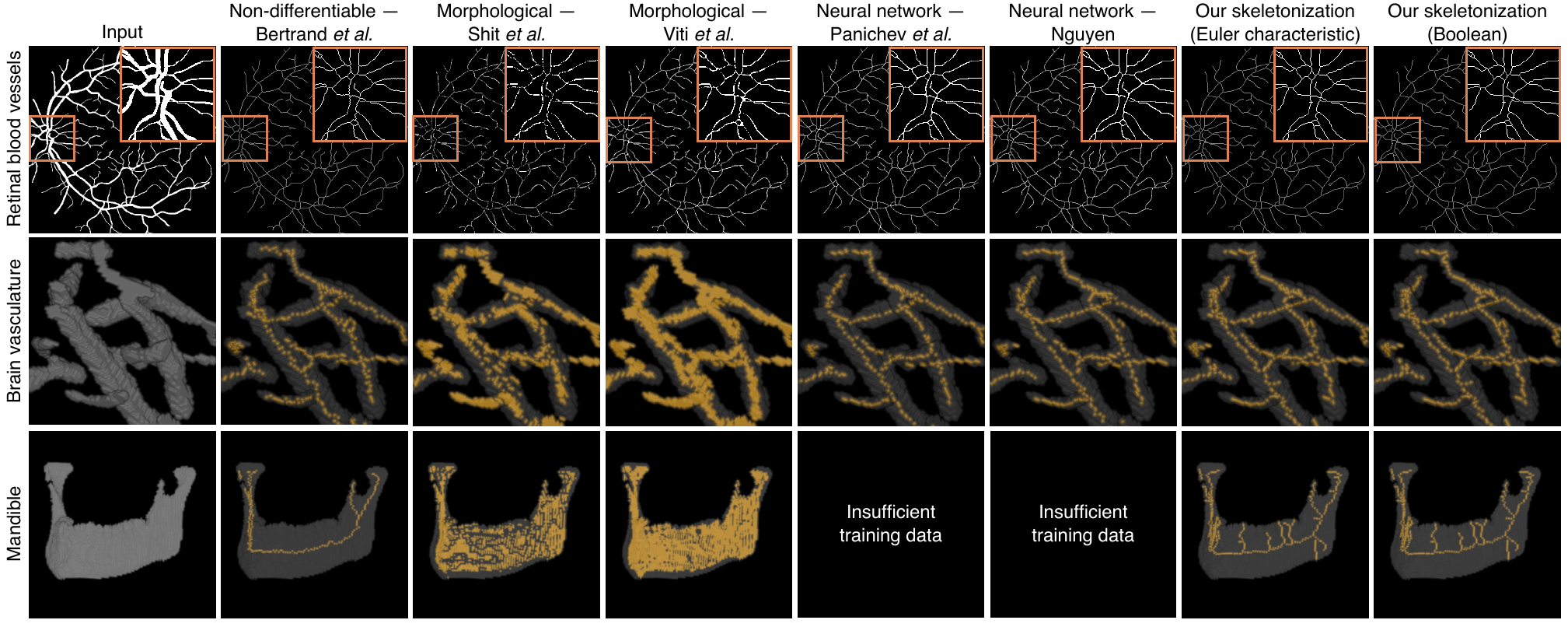}
\caption{The results of applying the seven tested skeletonization algorithm to representative samples of three diverse datasets. Of the six algorithms that are compatible with gradient-based optimization, only our two methods are able to extract a thin, topology-preserving skeleton, similar to the one obtained using the non-differential baseline. Additional samples are shown in the Supplementary Material.}
\label{fig:qualitative_results}
\end{figure*}

\subsection{Processing of continuous inputs}
\label{sec:continuous_inputs}

The previously introduced definitions for simple points and endpoints are only valid for binary images. However, in many applications the input is often a matrix of continuous values, such as probability maps output by a learning algorithm.
Simply rounding these inputs inhibits learning via backpropagation. We circumvent this issue by treating each point as a discrete, stochastic node modeled by a Bernoulli distribution, and use the reparametrization trick and a straight-through estimator to facilitate sampling and gradient estimation from it \cite{bengio2013estimating,jang2016categorical,maddison2017concrete}.

The reparametrization trick splits each node into a differentiable function, the raw grayscale input, and a fixed noise distribution \cite{huijben2022review,jang2016categorical,maddison2017concrete}. We can sample from each node via
\begin{equation}
X = \sigma \left({\frac{(\log\alpha + \beta L)}{\tau}}\right)\,,\,\alpha = \frac{\alpha_1}{(1 - \alpha_1)}\,, 
\end{equation}
where $\alpha_1\in(0,1)$ is the probability of the input being 1, $\sigma$ denotes the sigmoid function, $L$ is a sample from a Logistic distribution that is scaled by factor $\beta\in[0,\infty)$, and $\tau\in(0,\infty)$ is the Boltzmann temperature. Both $\beta$ and $\tau$ control the entropy of the distribution. Others have proposed gradually annealing these parameters as learning progresses or even updating them via backpropagation \cite{huijben2022review,jang2016categorical,maddison2017concrete}. In this work, we treat them as simple tunable hyperparameter.

Afterwards, we discretize the obtained samples using a straight-through gradient estimator \cite{bengio2013estimating,rosenblatt1957perceptron}. It returns the rounded binary value during the forward pass. Instead of using the zero gradient of the rounding operation during the backward pass, the modified chain rule is applied and the identity function is used as proxy gradient.

\subsection{Implementation in PyTorch}
\label{sec:implementation}

Our proposed algorithm consists exclusively of matrix additions, multiplications, convolutional operations, basic activation functions, and sampling from a uniform probability distribution. As such, it can easily be implemented in any major deep learning library and runs efficiently on graphics processing units. Our skeletonization module, which we make publicly available\footnote{\url{https://github.com/martinmenten/skeletonization-for-gradient-based-optimization}}, is implemented in PyTorch \cite{paszke2017automatic} and is fully integrated with its automatic differentiation engine.

\section{Experiments and results}

\begin{table*}[t!]
\centering
\caption{Quantitative comparison of the topological accuracy and run time of seven skeletonization algorithms on three datasets.}
\label{tab:benchmarking}
\small
\begin{tabular}{l l c c c c c}
\hline
Dataset & Skeletonization algorithm & \myhash\,points & $\beta_0$ error & $\beta_1$ error & $\beta_2$ error & Run time [ms] \\ \hline
\multirow{7}{*}{DRIVE} & Non-differentiable -- Bertrand \etal \cite{bertrand2014powerful} & 8316$\pm$618\hphantom{0} & 0$\pm$0 & 0$\pm$0 & - & - \\
& Morphological -- Shit \etal \cite{shit2021cldice} & 9926$\pm$667\hphantom{0} & 1156$\pm$197\hphantom{0} & 50$\pm$23 & - & 19$\pm$1\hphantom{0} \\
& Morphological -- Viti \etal \cite{viti2022coronary} & 11834$\pm$976\hphantom{00} & 266$\pm$62\hphantom{0} & 45$\pm$19 & - & 23$\pm$3\hphantom{0} \\
& Neural network -- Panichev \etal \cite{panichev2019u} & 10420$\pm$915\hphantom{00} & 6$\pm$3 & 13$\pm$11 & - & 14$\pm$1\hphantom{0} \\
& Neural network -- Nguyen \cite{nguyen2021u} & 10619$\pm$806\hphantom{00} & 10$\pm$5\hphantom{0} & 18$\pm$8\hphantom{0} & - & 117$\pm$1\hphantom{00} \\
\cline{2-7}
& Ours -- Euler characteristic & 8393$\pm$611\hphantom{0} & 0$\pm$0 & 0$\pm$0 & - & 101$\pm$3\hphantom{00} \\
& Ours -- Boolean & 8393$\pm$611\hphantom{0} & 0$\pm$0 & 0$\pm$0 & - & 540$\pm$2\hphantom{00} \\ \hline
\multirow{7}{*}{VesSAP} & Non-differentiable -- Bertrand \etal \cite{bertrand2014powerful} & 471$\pm$212 & 0$\pm$0 & 0$\pm$0 & 0$\pm$0 & - \\
& Morphological -- Shit \etal \cite{shit2021cldice} & 1914$\pm$809\hphantom{0} & 173$\pm$79\hphantom{0} & 3$\pm$5 & 0$\pm$1 & 20$\pm$1\hphantom{0} \\
& Morphological -- Viti \etal \cite{viti2022coronary} & 3783$\pm$1797 & 2$\pm$2 & 23$\pm$18 & 0$\pm$1 & 21$\pm$2\hphantom{0} \\
& Neural network -- Panichev \etal \cite{panichev2019u} & 423$\pm$182 & 64$\pm$38 & 3$\pm$5 & 0$\pm$1 & 16$\pm$1\hphantom{0} \\
& Neural network -- Nguyen \cite{nguyen2021u} & 422$\pm$189 & 33$\pm$19 & 4$\pm$6 & 0$\pm$1 & 115$\pm$1\hphantom{00} \\
\cline{2-7}
& Ours (Euler characteristic) & 540$\pm$245 & 0$\pm$1 & 0$\pm$1 & 0$\pm$1 & 100$\pm$2\hphantom{00} \\
& Ours (Boolean) & 540$\pm$243& 0$\pm$0 & 0$\pm$0 & 0$\pm$0 & 520$\pm$26\hphantom{0}  \\\hline
\multirow{7}{*}{Mandible} & Non-differentiable -- Bertrand \etal \cite{bertrand2014powerful} & 236$\pm$37\hphantom{0} & 0$\pm$0 & 0$\pm$0 & 0$\pm$0 & - \\
& Morphological -- Shit \etal \cite{shit2021cldice} & 2698$\pm$387\hphantom{0} & 131$\pm$24\hphantom{0} & 13$\pm$10 & 0$\pm$0 & 37$\pm$1\hphantom{0} \\
& Morphological -- Viti \etal \cite{viti2022coronary} & 4866$\pm$758\hphantom{0} & 1$\pm$1 & 81$\pm$26 & 0$\pm$0 & 42$\pm$1\hphantom{0} \\
& Neural network -- Panichev \etal \cite{panichev2019u} & \multicolumn{5}{c}{Insufficient training data} \\
& Neural network -- Nguyen \cite{nguyen2021u} &  \multicolumn{5}{c}{Insufficient training data} \\
\cline{2-7}
& Ours (Euler characteristic)& 409$\pm$67\hphantom{0} & 1$\pm$1 & 1$\pm$1 & 0$\pm$0 & 160$\pm$2\hphantom{00} \\
& Ours (Boolean) & 405$\pm$68\hphantom{0} & 0$\pm$0 & 0$\pm$0 & 0$\pm$0 & 1081$\pm$10\hphantom{00} \\ \hline
\end{tabular}
\end{table*}

Initially, we benchmark the performance of our skeletonization algorithm with regard to spatial and topological correctness, run time, and the ability to combine it with backpropagation (see Section \ref{sec:benchmarking_experiments}). Afterwards, we showcase the utility of our method by integrating it with two medical image processing pipelines: deep-learning-based blood vessel segmentation and multimodal registration of the mandible (see Section \ref{sec:application_experiments}).

For the experiments, we use three different datasets:
\begin{itemize}
\setlength\itemsep{0em}
\item the DRIVE dataset consisting of 40 two-dimensional retinal color fundus photographs and matching annotations of the visible blood vessels \cite{staal2004ridge},
\item the VesSAP dataset comprising 24 three-dimensional light-sheet microscopy images of murine brains after tissue clearing, staining, and labeling of the vascular network, which we split into 2,400 patches, \cite{todorov2020machine}.
\item an in-house dataset of 34 matched three-dimensional CT and MR images and manually extracted segmentation masks of the mandible.
\end{itemize}
Additional information about each dataset and our experimental setup can be found in the Supplementary Material.

\newpage
\subsection{Benchmarking experiments}
\label{sec:benchmarking_experiments}
\subsubsection{Spatial and topological accuracy}

We compare our two skeletonization algorithms with five baselines:

\begin{itemize}
\setlength\itemsep{0em}
\item a well-established non-differentiable skeletonization algorithm by Bertrand \etal \cite{bertrand2014powerful}, which has been implemented in the open-source DGtal library \cite{DGtal},
\item two morphological skeletonization algorithm based on repeated opening and erosion proposed by Shit \etal \cite{shit2021cldice} and Viti \etal \cite{viti2022coronary}, respectively,
\item two neural-network-based methods by Panichev \etal \cite{panichev2019u} and Nguyen \cite{nguyen2021u}, respectively, that each train a encoder-decoder network to output the skeleton of a binary input image.
\end{itemize}

On all three datasets, both of our proposed algorithms produce continuous, thin skeletons that agree well with the non-differentiable baseline (see Figure \ref{fig:qualitative_results}). When applying the morphological skeletonization algorithms, we observe that continuous blood vessels are split into many small objects along the medial axis. Similarly, the mandible is broken into small components that are positioned at the medial surface of the input structure. The neural-network-based algorithms also cannot preserve the topology of the vascular network when applied to data from the DRIVE and VesSAP datasets, and completely fail to converge during training on the small mandible dataset.

These observations are corroborated by quantitative measurements (see Table \ref{tab:benchmarking}). We assess the topological correctness of all skeletons by evaluating the error of the first three Betti numbers, $\beta_0$, $\beta_1$, and $\beta_2$. These measure the absolute difference of the number of objects, holes, and cavities, respectively, between the input structure and obtained skeleton. Our skeletonization algorithm based on the Boolean characterization of simple points preserves the exact topology of the base object in all cases as does the non-differentiable baseline. The skeletons obtained by the morphological and neural-network-based algorithms both contain topological errors in all three Betti numbers. Furthermore, their produced skeletons are often thicker than one voxel. This is reflected by the substantially larger number of points included in the results.

\subsubsection{Run time analysis}

Table \ref{tab:benchmarking} also lists the average run time of each skeletonization algorithm when processing images of varying sizes and dimensions. We report the duration of a single forward and backward pass through each skeletonization module as required during gradient-based optimization. All measurements were conducted using a workstation equipped with a Nvidia RTX A6000 GPU (Nvidia Corporation, Santa Clara, California, United States), 128 GB of random access memory, and a AMD Ryzen Threadripper 3970X 32-core central processing unit (Advanced Micro Devices, Inc., Santa Clara, California, United States). We find that our algorithms are slower than both morphological and neural-network-based methods. Still, all algorithms run in a second or less and are fast enough to be effectively employed for the applications described in the following (see Section \ref{sec:application_experiments}).

\subsubsection{Compatibility with gradient-based optimization}

\begin{figure}[h!]
\centering
\includegraphics[width=0.95\linewidth]{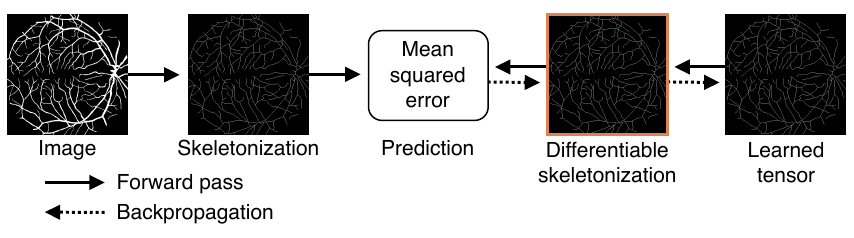}
\caption{Experiment to test the compatibility of our skeletonization algorithm (orange box) with backpropagation.}
\label{fig:naive_experiment}
\end{figure}

In order to make our skeletonization algorithms compatible with gradient-based optimization, logistic noise is added to the input during the reparametrization trick (see Section \ref{sec:continuous_inputs}). We demonstrate the efficiency of this approach and the importance of well-tuned noise parameters, $\beta$ and $\tau$, in a simple experiment (see Figure \ref{fig:naive_experiment}). Hereby, an input tensor is initialized with random values and passed through our skeletonization module. Using backpropagation, the tensor's values are learned so that its ultimate output resembles that of the ground truth skeleton. While several degenerate solutions that all yield the same skeleton, exist, we expect the learned tensor to ultimately resemble the skeleton itself.

\begin{figure}[h!]
\centering
\includegraphics[width=0.95\linewidth]{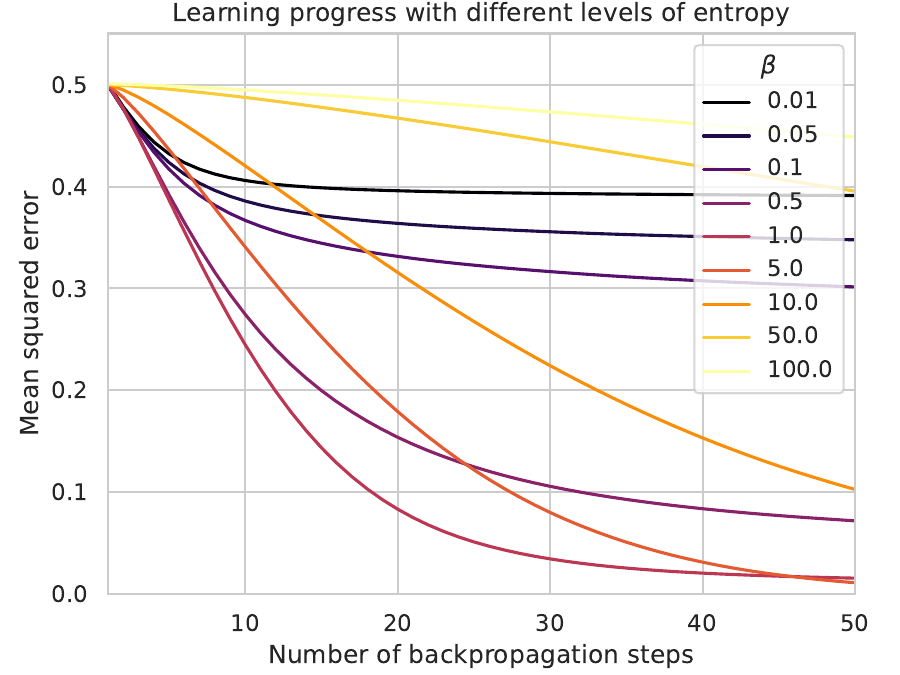}
\caption{Effect of the scale $\beta$ of the added logistic noise on the ability to propagate a gradient through our skeletonization module. Both very low entropy and very high entropy inhibit learning. Similar results can be found when varying $\tau$ (see Supplementary Material).}
\label{fig:learning_progress}
\end{figure}

\begin{figure}[h!]
\centering
\includegraphics[width=0.95\linewidth]{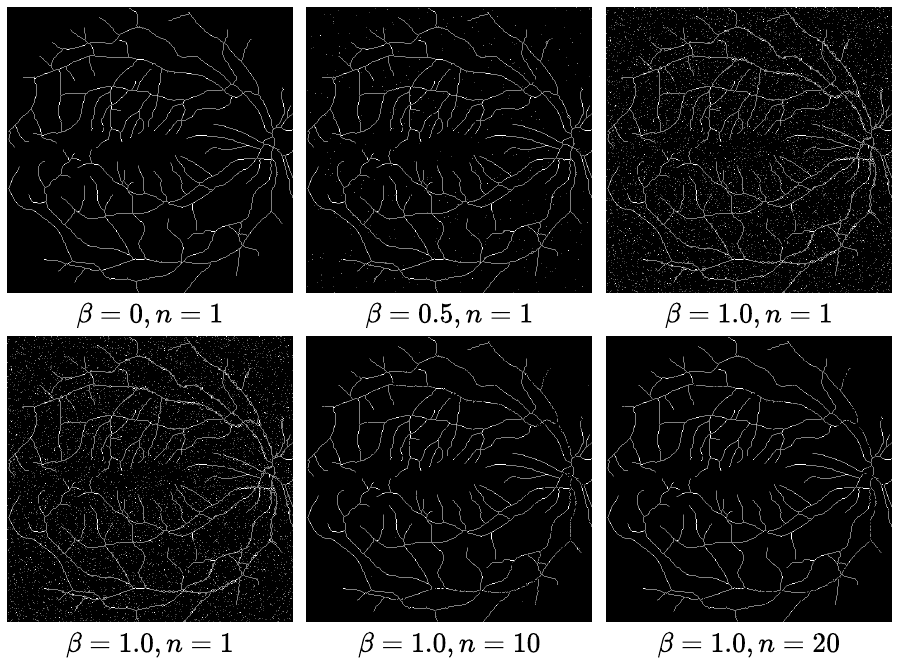}
\caption{Effect of scaling the added logistic noise ($\beta$) in our skeletonization algorithm. Repeated sampling ($n$) mostly recovers the true skeleton.}
\label{fig:effect_of_stochasticity}
\end{figure}

With very low levels of entropy, we observe that learning with our skeletonization module is very slow (see Figure \ref{fig:learning_progress}). Increasing the entropy results in single passes through the skeletonization to be less faithful to the geometry and topology of the true medial axis (see Figure \ref{fig:effect_of_stochasticity}). However, averaging over repeated samples mostly recovers the true skeleton and enables learning of the correct structure. At too high entropy, convergence slows down as the obtained skeleton is not sufficiently accurate anymore.

\subsection{Application experiments}
\label{sec:application_experiments}
\subsubsection{Topology-aware blood vessel segmentation}
\label{sec:vessel_segmentation}

\begin{figure}[h!]
\centering
\includegraphics[width=0.95\linewidth]{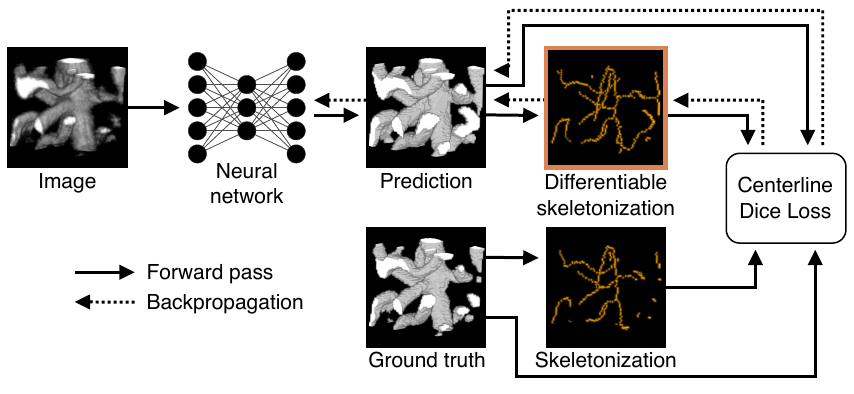}
\caption{Deep learning pipeline for training a vessel segmentation network using the centerline Dice loss \cite{shit2021cldice}. The loss formulation requires the use of a skeletonization algorithm that is compatible with backpropagation (orange box).}
\label{fig:vessel_segmentation_workflow}
\end{figure}

We explore the utility of our skeletonization methods by integrating them with a deep-learning-based segmentation algorithm for the VesSAP dataset (see Figure \ref{fig:vessel_segmentation_workflow}). The training of the basic U-Net incorporates the centerline Dice (clDice) loss function that encourages topology preservation across different vessel scales by comparing skeletons of the prediction and ground truth \cite{ronneberger2015u,shit2021cldice}. The loss formulation requires a skeletonization function that is compatible with backpropagation. In all cases, we tune the weighting factor $\lambda$, which balances the clDice loss with the standard Dice loss.

Using the clDice loss instead of a vanilla Dice loss slightly improves the topological agreement between prediction and ground truth as indicated by a lower error of the first three Betti numbers (see Table \ref{tab:vessel_segmentation_workflow}). Moreover, we find that using our skeletonization methods yield slightly better results than using a morphological skeletonization algorithm. Spatial accuracy, quantified by the Dice similarity coefficient (DSC), was nearly identical in all cases. We obtain similar findings when conducting the same experiment using the DRIVE dataset (see Supplementary Material).

\begin{table}[htbp]
\centering
\caption{Performance of the vessel segmentation network using either a standard Dice loss ('Without') or clDice loss with different skeletonization algorithms.}
\label{tab:vessel_segmentation_workflow}
\footnotesize
\begin{tabular}{l c c c c}
\hline
Skeletonization & DSC & $\beta_0$ error & $\beta_1$ error & $\beta_2$ error \\ \hline
Without & 0.85$\pm$0.01 & 5.1$\pm$0.8 & 3.1$\pm$0.1 & 0.8$\pm$0.3 \\
Morphological & 0.85$\pm$0.01 & 4.3$\pm$0.5 & 2.8$\pm$0.2 & 0.7$\pm$0.1 \\
Ours (Euler) & 0.86$\pm$0.01 & 3.5$\pm$0.2 & 2.7$\pm$0.1 & 0.4$\pm$0.1 \\
Ours (Boolean) & 0.86$\pm$0.01 & 3.7$\pm$0.3 & 2.8$\pm$0.1 & 0.5$\pm$0.2 \\
\hline
\end{tabular}
\end{table}

\subsubsection{Multimodal registration of the mandible in CT and MR images}
\label{sec:mandible_registration}

\begin{figure}[h!]
\centering
\includegraphics[width=0.95\linewidth]{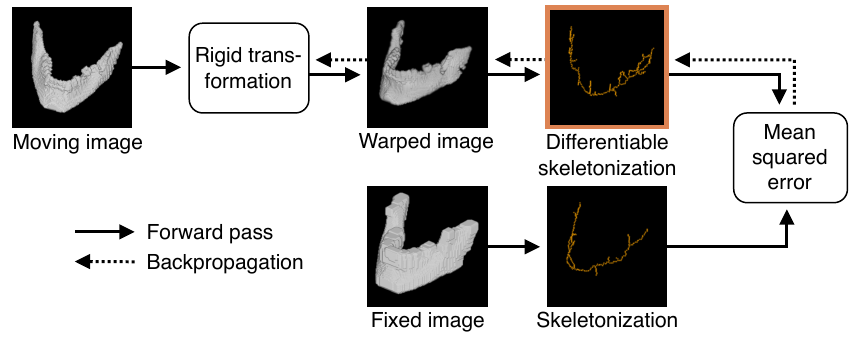}
\caption{Workflow for multimodal registration of the mandible. To compensate for the different size of the mandible in CT and MR images, the skeleton of both images are calculated (orange box) and registered instead.}
\label{fig:mandible_registration_workflow}
\end{figure}

Finally, we explore whether incorporating skeletonization can improve multimodal registration of the mandible (see Figure \ref{fig:mandible_registration_workflow}). This application is motivated by the fact that bones often appear larger in MR images than in CT images. When registering segmentation masks from both modalities, the smaller mask can be orientated flexibly inside the larger one. We propose extracting the skeleton of both structures and calculating their overlap as image distance function instead. We employ a conventional registration algorithm that optimizes the image distance with respect to the rigid transformation between both images using a gradient-based optimization method, thus requiring a compatible skeletonization method. We implement this application in AirLab \cite{Sandkuehler2018}, which uses Pytorch's autograd functionality to compute the gradient based on the objective function. This allows a seamless integration of our skeletonization module in a conventional registration algorithm.

We report the DSC, the Hausdorff distance (HD) and the average surface distance (ASD) between the fixed and warped segmentation as proxy measure for registration accuracy (see Table \ref{tab:mandible_registration_workflow}). Our findings show registering the images based on the skeleton of their segmentation map slightly improves the alignment of both structures.

\begin{table}[htbp]
\centering
\caption{Registration accuracy using two different loss functions: either the image distance is calculated using the full binary mask of the mandible ('Without'), or based on their skeletons obtained via one of three skeletonization algorithms.}
\label{tab:mandible_registration_workflow}
\footnotesize
\begin{tabular}{l c c c}
\hline
Skeletonization & DSC & HD [mm] & ASD [mm] \\ \hline
Without & 0.38$\pm$0.01 & 29.9$\pm$0.9 & 6.6$\pm$0.2 \\
Morphological & 0.32$\pm$0.01 & 29.2$\pm$1.1 & 6.7$\pm$0.2 \\
Ours (Euler) & 0.37$\pm$0.02 & 28.6$\pm$1.2 & 6.6$\pm$0.2\\
Ours (Boolean) & 0.37$\pm$0.01 & 28.0$\pm$1.1 & 6.5$\pm$0.2 \\
\hline
\end{tabular}
\end{table}

\section{Discussion and conclusion}

This work bridges the gap between classical skeletonization algorithms and the ability to integrate these with gradient-based optimization. We have introduced a three-dimensional skeletonization algorithm that is compatible with backpropagation, domain-agnostic and preserves an object’s topology. Our method combines a characterizations of simple points, a parallelization scheme for their efficient removal, and a strategy for discretization of non-binary inputs that are all compatible with gradient-based optimization. In benchmarking experiments, we have proved the superior spatial and topological accuracy of our method compared to morphological, and neural-network-based baselines.

Our algorithm consists exclusively of matrix additions, multiplications, convolutional operations, activation functions and sampling from basic probability distributions. Consequently, it can be implemented in any deep learning library and can be seamlessly integrated with diverse image processing pipelines that use gradient-based optimization. We showcase this utility by applying it in two realistic medical image processing applications: semantic segmentation of blood vessels with deep learning, and automated multimodal image registration. In both cases, we find that our skeletonization algorithms allows the incorporation of topological and geometric information within the respective optimization objective, leading to modest performance gains.

To our knowledge, this work introduces the first topology-preserving skeletonization algorithm for gradient-based optimization. Still, we discern that there may be other, potentially more effective, approaches to create such algorithms. We hope that this work can serve as a blueprint for others to further explore skeletonization. Building upon a rich body of literature on classical skeletonization algorithms, future work could further explore alternative strategies to identify simple points \cite{saha2016survey,shen2013skeleton}. Similarly, past works have extensively studied schemes to efficiently remove simple points in parallel of which some may be better suited for processing on graphics processing units. Finally, the endpoint condition used during skeletonization influences the properties of the created skeleton \cite{shen2013skeleton,zhou1999efficient}. In other applications skeletal surfaces may be preferable over a medial axis or a different trade-off between a geometric and topological skeleton may be chosen. Ultimately, we envision that our method may also be beneficial in many computer vision applications that have historically utilized skeletonization, but have since been increasingly solved using deep learning.

\section*{Acknowledgments}

We would like to thank Viktoria Thierauf for her help with the Mandible dataset. This work was funded by the Munich Center for Machine Learning.

{\small
\bibliographystyle{ieee_fullname}
\bibliography{bibliography}

\begin{thebibliography}{10}\itemsep=-1pt

\bibitem{DGtal}
Dgtal: Digital geometry tools and algorithms library.
\newblock \url{http://dgtal.org}.

\bibitem{tensorflow2015-whitepaper}
Mart\'{i}n Abadi et~al.
\newblock {TensorFlow}: Large-scale machine learning on heterogeneous systems,
  2015.

\bibitem{bengio2013estimating}
Yoshua Bengio et~al.
\newblock Estimating or propagating gradients through stochastic neurons for
  conditional computation.
\newblock {\em arXiv preprint arXiv:1308.3432}, 2013.

\bibitem{bertrand1996boolean}
Gilles Bertrand.
\newblock A boolean characterization of three-dimensional simple points.
\newblock {\em Pattern recognition letters}, 17(2):115--124, 1996.

\bibitem{bertrand1995three}
Gilles Bertrand et~al.
\newblock Three-dimensional thinning algorithm using subfields.
\newblock In {\em Vision Geometry III}, volume 2356, pages 113--124. SPIE,
  1995.

\bibitem{bertrand2014powerful}
Gilles Bertrand et~al.
\newblock Powerful parallel and symmetric 3d thinning schemes based on critical
  kernels.
\newblock {\em Journal of Mathematical Imaging and Vision}, 48:134--148, 2014.

\bibitem{blum1967transformation}
Harry Blum.
\newblock A transformation for extracting new descriptions of shape.
\newblock {\em Models for the perception of speech and visual form}, pages
  362--380, 1967.

\bibitem{borgefors1999computing}
Gunilla Borgefors et~al.
\newblock Computing skeletons in three dimensions.
\newblock {\em Pattern recognition}, 32(7):1225--1236, 1999.

\bibitem{demir2019skelneton}
Ilke Demir et~al.
\newblock Skelneton 2019: Dataset and challenge on deep learning for geometric
  shape understanding.
\newblock In {\em Proceedings of the IEEE/CVF Conference on Computer Vision and
  Pattern Recognition Workshops}, 2019.

\bibitem{ganin2016domain}
Yaroslav Ganin et~al.
\newblock Domain-adversarial training of neural networks.
\newblock {\em The journal of machine learning research}, 17(1):2096--2030,
  2016.

\bibitem{huijben2022review}
Iris~AM Huijben et~al.
\newblock A review of the gumbel-max trick and its extensions for discrete
  stochasticity in machine learning.
\newblock {\em IEEE Transactions on Pattern Analysis and Machine Intelligence},
  2022.

\bibitem{jang2016categorical}
Eric Jang et~al.
\newblock Categorical reparameterization with gumbel-softmax.
\newblock {\em ICLR}, 2017.

\bibitem{jerripothula2017object}
Koteswar~Rao Jerripothula et~al.
\newblock Object co-skeletonization with co-segmentation.
\newblock In {\em Proceedings of the IEEE Conference on Computer Vision and
  Pattern Recognition (CVPR)}, pages 3881--3889. IEEE, 2017.

\bibitem{kingma2014adam}
Diederik~P Kingma et~al.
\newblock Adam: A method for stochastic optimization.
\newblock {\em arXiv preprint arXiv:1412.6980}, 2014.

\bibitem{kong1989digital}
T~Yung Kong et~al.
\newblock Digital topology: Introduction and survey.
\newblock {\em Computer Vision, Graphics, and Image Processing},
  48(3):357--393, 1989.

\bibitem{lee1994building}
Ta-Chih Lee et~al.
\newblock Building skeleton models via 3-{D} medial surface axis thinning
  algorithms.
\newblock {\em CVGIP: Graphical Models and Image Processing}, 56(6):462--478,
  1994.

\bibitem{liu2020dynamic}
Jiang-Jiang Liu et~al.
\newblock Dynamic feature integration for simultaneous detection of salient
  object, edge, and skeleton.
\newblock {\em IEEE Transactions on Image Processing}, 29:8652--8667, 2020.

\bibitem{lobregt1980three}
Steven Lobregt et~al.
\newblock Three-dimensional skeletonization: principle and algorithm.
\newblock {\em IEEE Transactions on pattern analysis and machine intelligence},
  2(1):75--77, 1980.

\bibitem{ma1996fully}
C~Min Ma et~al.
\newblock A fully parallel 3{D} thinning algorithm and its applications.
\newblock {\em Computer vision and image understanding}, 64(3):420--433, 1996.

\bibitem{maddison2017concrete}
Chris~J. Maddison et~al.
\newblock The concrete distribution: A continuous relaxation of discrete random
  variables.
\newblock In {\em Proceedings of the International Conference on learning
  Representations}. International Conference on Learning Representations, 2017.

\bibitem{maragos1986morphological}
Petros Maragos et~al.
\newblock Morphological skeleton representation and coding of binary images.
\newblock {\em IEEE Transactions on Acoustics, Speech, and Signal Processing},
  34(5):1228--1244, 1986.

\bibitem{morgenthaler1981three}
David~G Morgenthaler.
\newblock Three-dimensional simple points: serial erosion, parallel thinning
  and skeletonization.
\newblock {\em TR-1005}, 1981.

\bibitem{morse1993multiscale}
Bryan~S Morse et~al.
\newblock Multiscale medial analysis of medical images.
\newblock In {\em Biennial International Conference on Information Processing
  in Medical Imaging}, pages 112--131. Springer, 1993.

\bibitem{nathan2019skeletonnet}
Sabari Nathan et~al.
\newblock Skeletonnet: Shape pixel to skeleton pixel.
\newblock In {\em Proceedings of the IEEE/CVF Conference on Computer Vision and
  Pattern Recognition Workshops}, pages 0--0, 2019.

\bibitem{nguyen2021u}
Nam~Hoang Nguyen.
\newblock U-net based skeletonization and bag of tricks.
\newblock In {\em Proceedings of the IEEE/CVF International Conference on
  Computer Vision}, pages 2105--2109, 2021.

\bibitem{palagyi2001sequential}
K{\'a}lm{\'a}n Pal{\'a}gyi et~al.
\newblock A sequential 3{D} thinning algorithm and its medical applications.
\newblock In {\em Biennial International Conference on Information Processing
  in Medical Imaging}, pages 409--415. Springer, 2001.

\bibitem{panichev2019u}
Oleg Panichev et~al.
\newblock U-net based convolutional neural network for skeleton extraction.
\newblock In {\em Proceedings of the IEEE/CVF Conference on Computer Vision and
  Pattern Recognition Workshops}, 2019.

\bibitem{paszke2017automatic}
Adam Paszke et~al.
\newblock Automatic differentiation in pytorch.
\newblock {\em NIPS 2017 Workshop Autodiff}, 2017.

\bibitem{ronneberger2015u}
Olaf Ronneberger et~al.
\newblock U-net: Convolutional networks for biomedical image segmentation.
\newblock In {\em Medical Image Computing and Computer-Assisted
  Intervention--MICCAI 2015: 18th International Conference, Munich, Germany,
  October 5-9, 2015, Proceedings, Part III 18}, pages 234--241. Springer, 2015.

\bibitem{rosenblatt1957perceptron}
Frank Rosenblatt.
\newblock {\em The perceptron, a perceiving and recognizing automaton}.
\newblock Cornell Aeronautical Laboratory, 1957.

\bibitem{rosenfeld1976digital}
Azriel Rosenfeld et~al.
\newblock {\em Digital picture processing}.
\newblock Academic press, 1976.

\bibitem{saha1997new}
Punam~K Saha et~al.
\newblock A new shape preserving parallel thinning algorithm for {3D} digital
  images.
\newblock {\em Pattern recognition}, 30(12):1939--1955, 1997.

\bibitem{saha2016survey}
Punam~K Saha et~al.
\newblock A survey on skeletonization algorithms and their applications.
\newblock {\em Pattern recognition letters}, 76:3--12, 2016.

\bibitem{Sandkuehler2018}
Robin Sandk{\"u}hler et~al.
\newblock Airlab: autograd image registration laboratory.
\newblock {\em arXiv preprint arXiv:1806.09907}, 2018.

\bibitem{shaked1998pruning}
Doron Shaked et~al.
\newblock Pruning medial axes.
\newblock {\em Computer vision and image understanding}, 69(2):156--169, 1998.

\bibitem{shen2013skeleton}
Wei Shen et~al.
\newblock Skeleton pruning as trade-off between skeleton simplicity and
  reconstruction error.
\newblock {\em Science China Information Sciences}, 56:1--14, 2013.

\bibitem{shen2016object}
Wei Shen et~al.
\newblock Object skeleton extraction in natural images by fusing
  scale-associated deep side outputs.
\newblock In {\em Proceedings of the IEEE Conference on Computer Vision and
  Pattern Recognition}, pages 222--230, 2016.

\bibitem{shen2017deepskeleton}
Wei Shen et~al.
\newblock Deepskeleton: Learning multi-task scale-associated deep side outputs
  for object skeleton extraction in natural images.
\newblock {\em IEEE Transactions on Image Processing}, 26(11):5298--5311, 2017.

\bibitem{shit2021cldice}
Suprosanna Shit et~al.
\newblock cldice-a novel topology-preserving loss function for tubular
  structure segmentation.
\newblock In {\em Proceedings of the IEEE/CVF Conference on Computer Vision and
  Pattern Recognition}, pages 16560--16569, 2021.

\bibitem{staal2004ridge}
Joes Staal et~al.
\newblock Ridge-based vessel segmentation in color images of the retina.
\newblock {\em IEEE transactions on medical imaging}, 23(4):501--509, 2004.

\bibitem{thibault2000terrain}
David Thibault et~al.
\newblock Terrain reconstruction from contours by skeleton construction.
\newblock {\em GeoInformatica}, 4:349--373, 2000.

\bibitem{todorov2020machine}
Mihail~Ivilinov Todorov et~al.
\newblock Machine learning analysis of whole mouse brain vasculature.
\newblock {\em Nature methods}, 17(4):442--449, 2020.

\bibitem{tsao1981parallel}
YF Tsao et~al.
\newblock A parallel thinning algorithm for 3-{D} pictures.
\newblock {\em Computer graphics and image processing}, 17(4):315--331, 1981.

\bibitem{vazquez2013virtual}
David Vazquez et~al.
\newblock Virtual and real world adaptation for pedestrian detection.
\newblock {\em IEEE transactions on pattern analysis and machine intelligence},
  36(4):797--809, 2013.

\bibitem{viti2022coronary}
Mario Viti et~al.
\newblock Coronary artery centerline tracking with the morphological skeleton
  loss.
\newblock In {\em Proc. ICIP}, pages 2741--2745, 2022.

\bibitem{zhao2007preprocessing}
Feng Zhao et~al.
\newblock Preprocessing and postprocessing for skeleton-based fingerprint
  minutiae extraction.
\newblock {\em Pattern Recognition}, 40(4):1270--1281, 2007.

\bibitem{zhou1999efficient}
Yong Zhou et~al.
\newblock Efficient skeletonization of volumetric objects.
\newblock {\em IEEE Transactions on visualization and computer graphics},
  5(3):196--209, 1999.

\end{thebibliography}
}

\clearpage
\section*{Supplementary Material}
\subsection*{Used datasets}

\noindent\textbf{DRIVE} The widely used DRIVE dataset consists of 40 two-dimensional retinal color fundus photographs and matching annotations of the visible blood vessels \cite{staal2004ridge}. We normalize the images to an intensity range between 0 and 1 and crop them to a size of $512 \times 512$ pixel. Then, we divide the dataset into training, validation and testing splits with a ratio of 60\% to 20\% to 20\%.

\noindent\textbf{VesSAP} The VesSAP dataset contains 24 three-dimensional light-sheet microscopy images of murine brains after tissue clearing, staining, and labeling of the vascular network. It has been made publicly available and has been extensively described by Todorov \etal \cite{todorov2020machine}. We split the $500 \times 500 \times 50$ voxel large images into non-overlapping patches of size $50 \times 50 \times 50$. We remove the patches that only contain background. Finally, we split the remaining ones into a training, validation and testing partition with a ratio of 80\% to 10\% to 10\%, while ensuring a subject-wise split.

\noindent\textbf{Mandible} The mandible dataset consists of 34 matched CT and MR images of the lower head and neck. In all images the mandible bone was outlined by a clinical expert. We resample all images to a resolution of $0.25 \times 0.25 \times 0.25$ cm$^3$ and subsequently remove all smaller cavities of the segmentation mask by alternatingly applying dilation and erosion operations. For the benchmarking experiments of the skeletonization algorithm we exclusively use the CT images (cf. Section 4.1 of the main paper). For the multimodal registration workflow that incorporates our skeletonization module we use the matched image pairs (cf. Section 4.3 of the main paper).

\subsection*{Neural network architecture and training}

This work uses neural networks either for explicit skeletonization (cf. Section 4.1 of the main paper) or for vessel segmentation (cf. Section 4.2 of the main paper). In the first case networks are provided with a binary mask and asked to provide the ground truth skeleton while being evaluated using the Dice loss. In the second case the neural network is provided images from either the DRIVE or VesSAP dataset and trained to output a blood vessel segmentation map. Hereby, we use the topology-preserving clDice loss in combination with various skeletonization algorithms \cite{shit2021cldice}.

The two neural-network-based skeletonization methods are implemented according to the works and accompanying public software code by Panichev \etal \cite{panichev2019u} and Nguyen \cite{nguyen2021u}, respectively. In order to facilitate processing of volumetric images we replace all two-dimensional operations, such as convolutions, pooling and normalization network layers, with their three-dimensional equivalents.

The segmentation neural networks follow a basic U-Net architecture with four downsampling and four upsampling blocks with skip connections \cite{ronneberger2015u}. Each block consists of two sequences of convolutional layer (either two- or three-dimensional convolutions, kernel size: 3, same padding), instance normalization layer and leaky-ReLU non-linearity (slope: 0.01). Downsampling is achieved by using a stride of 2 in the second convolutional layer of each block. At each downsampling step, the number of feature maps is also doubled and copied to the skip connection. Upsampling is achieved via a transposed convolution with a kernel size of 2 and stride of 2. After upsampling, the respective skip connection is concatenated with the main feature map. Network weights are optimized using the ADAM optimizer with a learning rate of $10^{-4}$ \cite{kingma2014adam}. In experiments using the DRIVE dataset, networks are trained with a batch size of 2 for 1,000 epochs. We apply random shifts ($\pm10\%$) and rotations ($\pm45^\circ$) as data augmentation. For the VesSAP dataset, networks are trained with a batch size of 16 for 200 epochs. In each case, we pick the best performing network based on the validation dataset before reporting the results on the test dataset. All experiments are repeated three times using different random seeds.

\subsection*{Additional qualitative skeletonization results}

\begin{figure*}[htbp]
\centering
\includegraphics[width=0.95\linewidth]{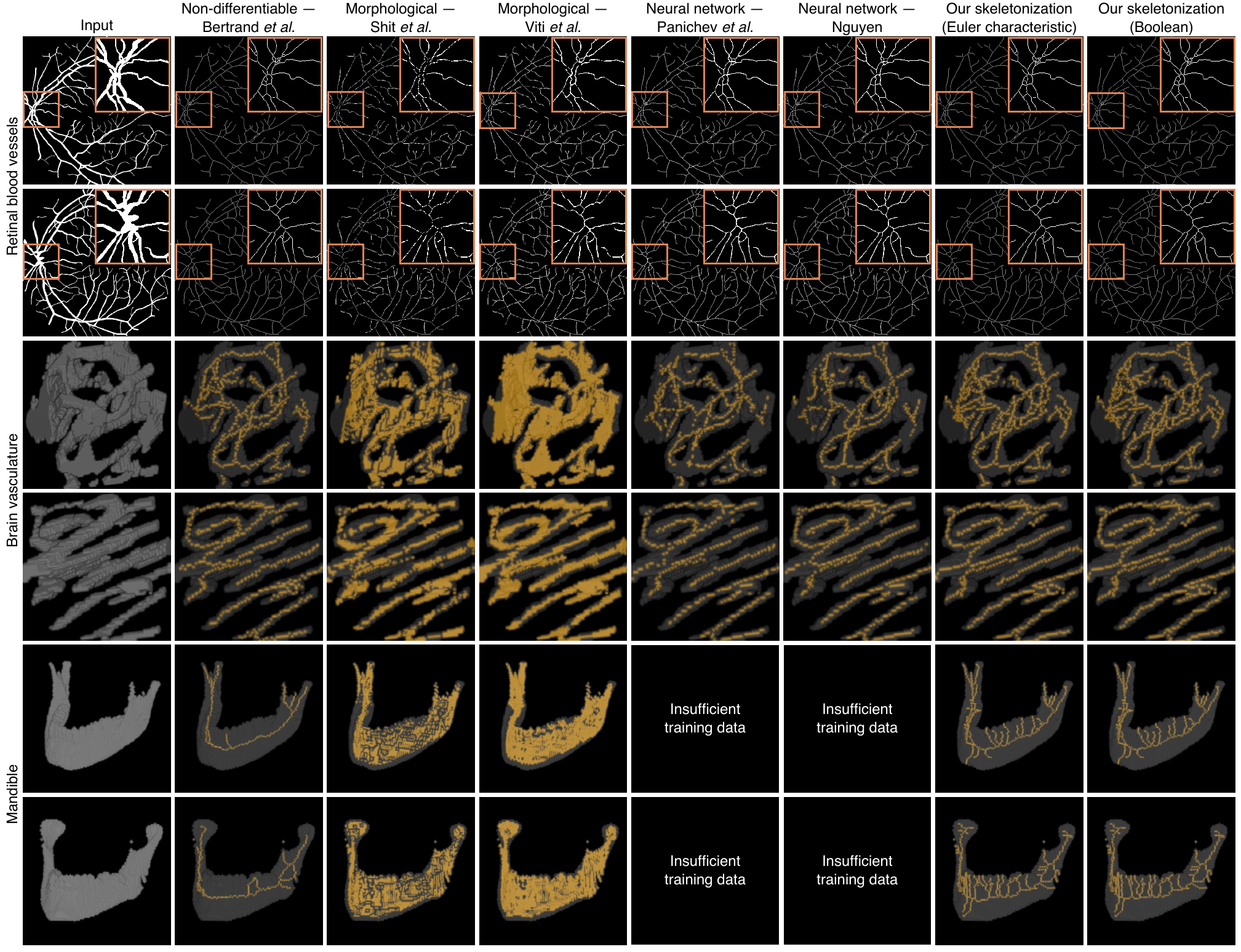}
\caption{Additional results of applying the seven tested skeletonization algorithm to representative samples of three diverse datasets. Of the six algorithms that are compatible with gradient-based optimization, only our two methods are able to extract a thin, topology-preserving skeleton, similar to the one obtained using the non-differential baseline.}
\label{fig:qualitative_results_sup}
\end{figure*}

Figure \ref{fig:qualitative_results_sup} presents additional representative results of applying the five skeletonization algorithms to the three datasets (cf. Section 4.1 of the main paper).

\subsection*{Experiments using the SkelNetOn dataset}

The SkelNetOn dataset was published in the scope of the  Deep Learning for Geometric Shape Understanding workshop held in conjunction with the 2019 IEEE/CVF Conference on Computer Vision and Pattern Recognition \cite{demir2019skelneton}. It consists of binary images depicting a range of stylized objects and their corresponding skeletons. Compared to the complex topologies of biological structures, the dataset exclusively features closed two-dimensional surfaces.

\begin{figure*}[htbp]
\centering
\includegraphics[width=0.95\linewidth]{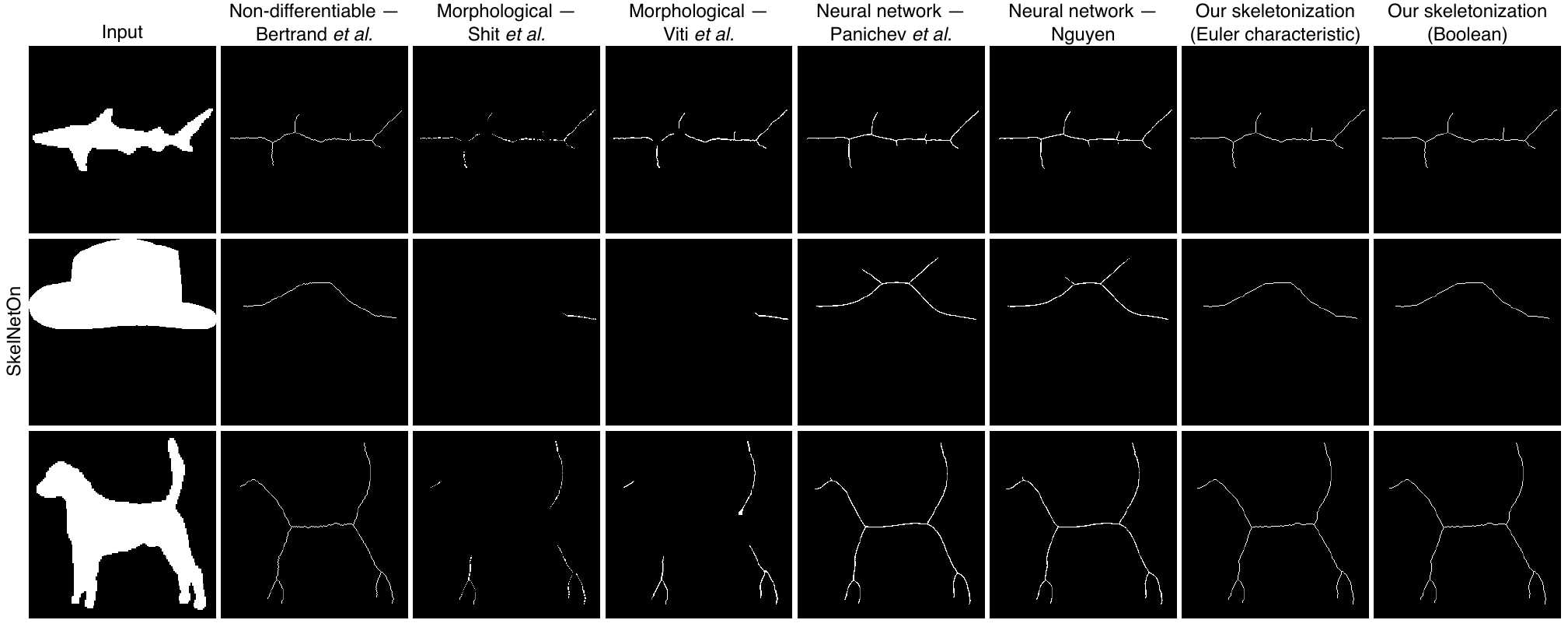}
\caption{Qualitative results of applying the seven tested skeletonization algorithm to representative samples of SkelNetOn dataset.}
\label{fig:qualitative_results_skelneton}
\end{figure*}

We repeat the same benchmarking experiments as described in Section 4.1 of the main paper using the SkelNetOn dataset. At the time of our study the challenge's public leaderboard had been taken offline, so that we could not use the official test split to benchmark our skeletonization algorithms. Instead, we split the training dataset into a training, validation and testing partition. We observe the same characteristic behavior of all skeletonization algorithms as in the experiments with the other three datasets (see Figure \ref{fig:qualitative_results_skelneton}). Both morphological baseline algorithms introduce breaks along the skeleton and in some cases omit large parts of the medial axis. The neural-network-based solutions also alter the topology of the object, whereas our skeletonization algorithms result in a topologically correct, thin skeleton. This is also reflected in the quantitative measurements reported in Table \ref{tab:benchmarking_skelneton}.

\begin{table*}[t!]
\centering
\caption{Quantitative comparison of the topological accuracy of seven skeletonization algorithms on the SkelNetOn dataset.}
\label{tab:benchmarking_skelneton}
\small
\begin{tabular}{l l c c c c c }
\hline
Dataset & Skeletonization algorithm & \myhash\,points & $\beta_0$ error & $\beta_1$ error & $\beta_2$ error & Run time [ms] \\ \hline
\multirow{7}{*}{SkelNetOn} & Non-differentiable -- Bertrand \etal \cite{bertrand2014powerful} & 355$\pm$206 & 0$\pm$0 & 0$\pm$0 & - & - \\
& Morphological -- Shit \etal \cite{shit2021cldice} & 158$\pm$150 & 40$\pm$30 & 0$\pm$1 & - & 23$\pm$2\hphantom{0} \\
& Morphological -- Viti \etal \cite{viti2022coronary} & 355$\pm$295 & 5$\pm$5 & 0$\pm$1 & - & 26$\pm$2\hphantom{0} \\
& Neural network -- Panichev \etal \cite{panichev2019u} & 524$\pm$247 & 2$\pm$2 & 0$\pm$1 & - & 30$\pm$1\hphantom{0} \\
& Neural network -- Nguyen \cite{nguyen2021u} & 494$\pm$236 & 2$\pm$3 & 0$\pm$1 & - & 160$\pm$2\hphantom{00} \\
\cline{2-7}
& Ours -- Euler characteristic & 406$\pm$241 & 0$\pm$0 & 0$\pm$0 & - & 189$\pm$3\hphantom{00} \\
& Ours -- Boolean & 406$\pm$241 & 0$\pm$0 & 0$\pm$0 & - & 948$\pm$5\hphantom{00} \\ \hline
\end{tabular}
\end{table*}

\subsection*{Effect of Boltzmann temperature on learning with the differentiable skeletonization module}

\begin{figure}[htbp]
\centering
\includegraphics[width=0.95\linewidth]{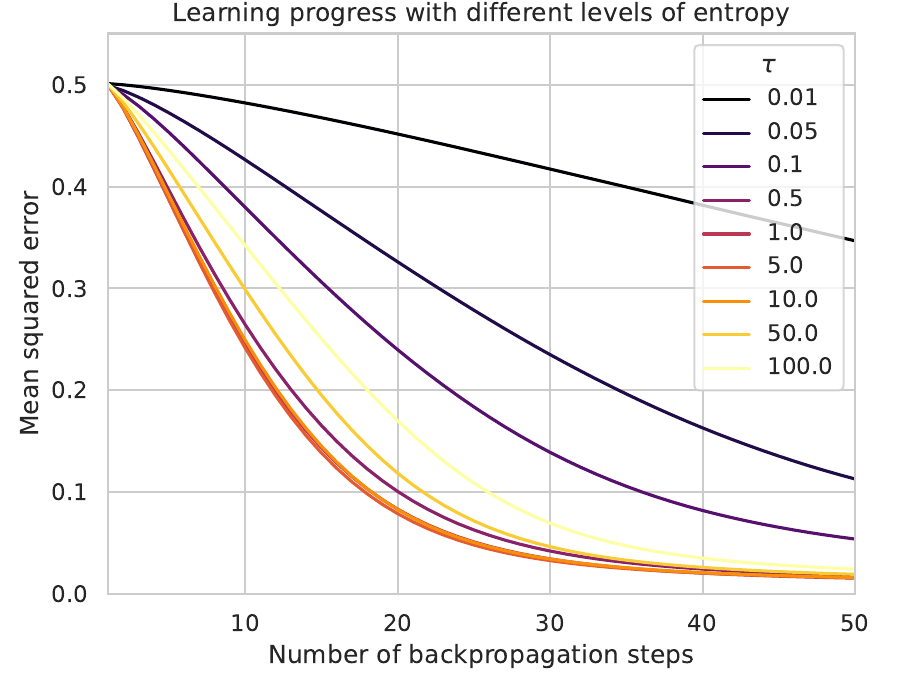}
\caption{Effect of the Boltzmann temperature $\tau$ on the ability to propagate a gradient through our skeletonization module. Both very low entropy and very high entropy inhibit learning.}
\label{fig:learning_progress_sup}
\end{figure}

The entropy of the stochastic discretization can be controlled by varying either the scale of the noise $\beta$ or the Boltzmann temperature $\tau$ (cf. Equation 11 of the main paper). We have also conducted the simple experiment presented in Figure 6 of the main paper while varying $\tau$ instead of $\beta$. Hereby, an input tensor is initialized with random values and passed through our skeletonization module. Using backpropagation, the tensor's values are learned so that its ultimate output resembles that of the ground truth skeleton. Analogously to our our results with varying noise scales (cf. Figure 7 of the main paper), we find that both a too low and too high Boltzmann temperature inhibit efficient learning with our skeletonization module (see Figure \ref{fig:learning_progress_sup}). Empirically, we find that it suffices to tune either the noise scale or Boltzmann temperature and proceed to tune $\beta$ throughout all other presented experiments.

\subsection*{Vessel segmentation in the DRIVE dataset}

As described above and in Section 4.2 of the main paper, we integrated our skeletonization modules with a neural network that learns to segment blood vessels in either the VesSAP or DRIVE dataset. The results on the two-dimensional DRIVE dataset are shown in Table \ref{tab:vessel_segmentation_sup}. Similar to the results for the VesSAP dataset (see Table 2 of the main paper), we find that using the clDice loss instead of a vanilla Dice loss slightly improves the topological agreement between prediction and ground truth as indicated by a lower error of the first two Betti numbers ($\beta_2$ indicating the difference in the number of cavities is always 0 in two dimensions). Moreover, we find that using our skeletonization methods yield slightly better results than using a morphological skeletonization algorithm. Spatial accuracy, quantified by the Dice similarity coefficient (DSC), is nearly identical in all cases.

\begin{table}[htbp]
\centering
\caption{Performance of the vessel segmentation network using either a standard Dice loss ('Without') or clDice loss with one of three skeletonization algorithms.}
\label{tab:vessel_segmentation_sup}
\footnotesize
\begin{tabular}{l c c c}
\hline
Skeletonization & DSC & $\beta_0$ error & $\beta_1$ error \\ \hline
Without & 0.79$\pm$0.01 & 135.6$\pm$5.7 & 23.3$\pm$2.1 \\
Morphological \cite{maragos1986morphological} & 0.79$\pm$0.01 & 58.2$\pm$2.1 & 21.3$\pm$1.2 \\
Ours (Euler) & 0.79$\pm$0.01 & 58.3$\pm$18.6 & 18.6$\pm$0.5 \\
Ours (Boolean) & 0.79$\pm$0.01 & 49.6$\pm$8.5 & 20.6$\pm$2.1 \\
\hline
\end{tabular}
\end{table}

\end{document}